\documentclass[runningheads]{llncs}

\usepackage{graphicx}
\usepackage{amssymb}
\usepackage{tabularx}
\usepackage{amsmath}
\usepackage{multirow}
\usepackage{cleveref}
\usepackage{booktabs}
\usepackage{longtable}
\usepackage{svg}
\usepackage{url}
\usepackage{float}
\usepackage{xspace}
\usepackage{tabularx}
\usepackage{array}
\usepackage{tabularx}
\usepackage[ruled,vlined]{algorithm2e}
\usepackage{pifont}
\usepackage{marvosym}

\usepackage{subcaption}
\usepackage{makecell}
\usepackage{fontawesome}

\newcommand{\questionone}{boundary discontinuity\xspace}
\newcommand{\questiontwo}{trajectory deformation\xspace}
\newcommand{\questionthree}{uneven pixel density\xspace}
\newcommand{\pointone}{adaptive reprojection\xspace}
\newcommand{\pointtwo}{great-circle trajectory adjustment\xspace}
\newcommand{\pointthree}{spherical search region tracking\xspace}

\newcommand{\Ours}{SphereDrag\xspace}
\newcommand{\OurBench}{PanoBench\xspace}
\newcommand{\pointones}{AR\xspace}
\newcommand{\pointtwos}{GCTA\xspace}
\newcommand{\pointthrees}{SSRT\xspace}  
\newcommand{\imgsize}{$1024 \times 512$\xspace}

\begin{document}
\title{\Ours: Spherical Geometry-Aware Panoramic Image Editing}
\author{Zhiao Feng\inst{1} \and
Xuewei Li\inst{1} \textsuperscript{(\Letter)} \and
Junjie Yang\inst{1} \and
Jingchao Li\inst{1} \and
Yuxin Peng\inst{2} \and
Xi Li \inst{3}
}
\authorrunning{Z. Feng et al.}
\institute{School of Electronic and Information Engineering, Shanghai DianJi University, Shanghai 201306, China \\
\email{xueweili@sdju.edu.cn}
\and
Department of Sports Science, Zhejiang University, Hangzhou 310058, China \and
College of Computer Science and Technology, Zhejiang University, Hangzhou 310058, China \\
}
\maketitle
\begin{abstract}
Image editing has made great progress on planar images, but panoramic image editing remains underexplored.
Due to their spherical geometry and projection distortions, panoramic images present three key challenges: \questionone, \questiontwo and \questionthree.
To tackle these issues, we propose \textbf{\Ours}, a novel panoramic editing framework utilizing spherical geometry knowledge for accurate and controllable editing.
% and distortion Reduction.
Specifically, \pointone(\pointones) uses adaptive spherical rotation to deal with discontinuity; \pointtwo(\pointtwos) tracks the movement trajectory more accurate; \pointthree(\pointthrees)  adaptively scales the search range based on spherical location to address uneven pixel density.
Also, we construct \OurBench, a panoramic editing benchmark, including complex editing tasks involving multiple objects and diverse styles, which provides a standardized evaluation framework. 
Experiments show that \Ours gains a considerable improvement compared with existing methods in geometric consistency and image quality, achieving up to 10.5\% relative improvement. 

\end{abstract}

\section{Introduction}
\begin{figure}[tb]
    \centering
    \includegraphics[width=\textwidth]{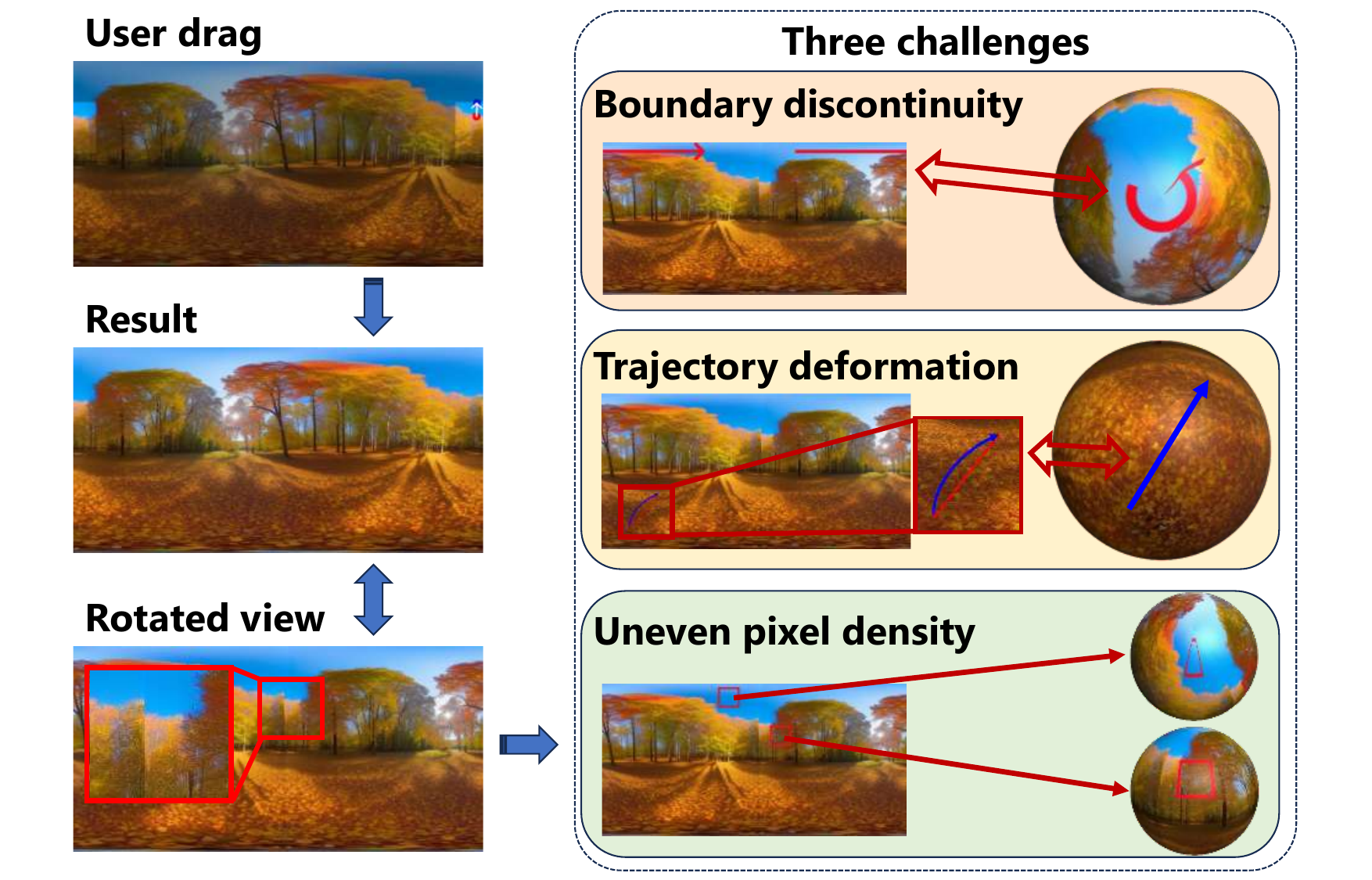}
    \caption{Illustration of challenges in panoramic image editing.
    (Upper right) The panoramic image may divide movement trajectory into two parts, located near the left and right boundaries, respectively. 
    (Middle right) Straight lines in the panoramic image do not correspond to great-circle paths on the sphere, leading to trajectory deviations. 
    (Lower right) The same region in the panoramic image corresponds to unequal solid angles at different latitudes, causing non-uniform tracking across the sphere.}
    \label{fig:fig1}
\end{figure}
Image generation is one of the current research hotspots. 
Many works~\cite{mou2024t2i,zhang2023adding,zheng2022entropy,jiang2025energy} have made significant progress in controllable high-quality planar image generation, primarily by fine-tuning pre-trained large-scale diffusion models to fit various application scenarios. 
Numerous studies~\cite{shi2024dragdiffusion,cui2024stabledrag,liu2024drag} have also been proposed for image editing, enabling highly controllable modifications to existing images based on new requirements. 
To gain a more comprehensive understanding of the scene, spherical panoramic images, also known as omnidirectional panoramic images or 360$^{\circ}$ panoramic images, are widely used in autonomous driving~\cite{de2018eliminating,li2023sgat4pass}, and virtual reality, etc. 

However, panoramic image editing is relatively underexplored.
Compared with normal plane images, panoramic images have three unique challenges caused by spherical geometry and image distortion that are appeared during projection process (e.g., Equirectangular Projection, ERP). 
First, \questionone occurs when the trajectories split at the left and right boundaries of the image. 
Second, \questiontwo arises because the straight trajectories on the sphere do not always correspond to straight trajectories in the panoramic images. 
Third, \questionthree results from the latitude-based distortion in panoramic images, which means the equal-size regions in the panoramic image correspond to different sizes on the sphere. 
These challenges are illustrated in~\cref{fig:fig1} in detail.
How can we design the image editing model to deal with these challenges above and utilize the characteristics of spherical images to realize better panoramic image editing?

Faced with these challenges, we propose the \Ours framework, which aims to achieve accurate and controllable panoramic point-interactive image editing by leveraging spherical knowledge.
We incorporate the characteristics of spherical panoramic images into the editing process. 
To address \questionone, we propose \pointone (\pointones), reprojecting the input panoramic images, that provides a more intuitive and editable representation. 
Also, we propose \pointtwo (\pointtwos), incorporating spherical knowledge to ease \questiontwo.
Additionally, to address \questionthree, we propose \pointthree (\pointthrees), which adaptively adjusts the search range based on spherical location information to mitigate non-uniform pixel density.
For a fair and comprehensive image editing evaluation, we introduce a new spherical point-interactive image editing benchmark \textbf{\OurBench}, including a lot of scenes with multiple objects and diverse styles. 
It not only includes basic single-object editing tasks but also supports challenging scenarios where the interplay of multiple objects and stylistic transformations must be handled simultaneously.  
Experiments show that \Ours considerably outperforms existing methods in terms of geometric consistency and image quality. 
At a 30° Field of View (FOV) evaluation setting, a 10.5\% relative improvement in IF is gained over the baseline, demonstrating the high-quality editing of \Ours. 
\Ours also achieves considerable reductions in both FID and sFID, indicating its advantages in both geometric accuracy and image quality.
Our contributions are summarized as follows:
\begin{itemize}
    \item We propose \Ours, a novel panoramic image editing framework that supports point-interactive editing by leveraging \pointones to address \questionone, \pointtwo to tackle \pointtwos, and \pointthrees to deal with \questionthree.
    \item We construct a spherical editing benchmark, \OurBench, which includes complex scenes with multiple objects and multiple styles, providing a standardized evaluation for panoramic image editing.
    \item Experiments show that \Ours considerably outperforms existing methods in terms of geometric consistency and image quality. 
\end{itemize}

\section{Related Work}
\label{relatedwork}

The two most related fields of our work are image editing and spherical panoramic image generation, which are discussed in \cref{imageediting} and \cref{sphimagegen}, respectively. 

\subsection{Image Editing}
\label{imageediting}
Image generation is a popular topic in image generation because of its user-friendly interaction. 
Early image editing methods are based on Generative Adversarial Networks (GANs)~\cite{goodfellow2020generative,karras2019style}, which made great progress in image generation but struggled with training instability and mode collapse~\cite{mescheder2018training,arjovsky2017wasserstein}.
These limitations make researchers explore more advanced generation models.
Diffusion models~\cite{ho2020denoising} become popular because of its ability to generate high-quality and highly controllable images through iterative denoising, and achieve remarkable improvements in both fidelity and diversity~\cite{saharia2022photorealistic,ramesh2022hierarchical}. 

Recently, many works focus on sophisticated image editing capabilities~\cite{kawar2023imagic,brooks2023instructpix2pix}. 
Early ones integrate natural language with generation models to enable text-guided editing~\cite{patashnik2021styleclip,gal2022stylegan}, but these methods suffer from insufficient precision due to inherent textual ambiguity \cite{zhang2023adding}, limiting their effectiveness for fine-grained adjustments. 
To deal with these challenges, researchers develop more precise point-driven editing methods based on interactive drag operations. 
DragGAN~\cite{pan2023drag} pioneered this approach by enabling intuitive local structure adjustments with direct user control. 
Building upon this foundation, subsequent works incorporate diffusion framework \cite{zheng2023layoutdiffusion}, leading to DragDiffusion~\cite{shi2024dragdiffusion}, which leverages diffusion models' advantages in maintaining global consistency while enabling local edits. 
DragNoise~\cite{liu2024drag} further refine this paradigm through innovative noise optimization strategies that better balance local control with global coherence. StableDrag~\cite{cui2024stabledrag} introduces a robust framework for point-based image editing by enhancing DragGAN and DragDiffusion with a discriminative point tracking method and a confidence-based latent enhancement strategy.

The methods listed above all focus on normal plane image editing. 
To address the special challenges of panoramic image editing, \Ours deals with the three special issues, and extends point-driven editing strategy to the panoramic image editing task.

\subsection{Spherical Panoramic Image Generation}
\label{sphimagegen}
In the contemporary panoramic image generation field, research methodologies are generally categorized into two main paradigms: GAN-based models and diffusion-based models. 

In the domain of GAN-based~\cite{goodfellow2020generative,karras2019style} approaches, COCO-GAN~\cite{lin2019coco} introduces a coordinate-conditional framework that generates images in a divided manner, using spatial coordinates as a guiding signal for the generator to progressively synthesize local regions. 
Its discriminator evaluates global consistency, local appearance, and edge continuity, ensuring seamless output and enabling prediction beyond training dimensions via cylindrical coordinates.
PanoGAN~\cite{wu2022cross} utilizes an adversarial feedback mechanism along with a cross-view generation framework, achieving a breakthrough in the semantic matching between aerial perspective and ground-level panoramic images. 
By using a dual-branch discriminator and repeated optimization, it can ensure the feature matching at the pixel level and maintain the spatial consistency within multimodal data. 
Similarly, SphericGAN~\cite{chen2022sphericgan} adopts a semi-supervised hyperspherical latent space modeling strategy, aligning real and generated data clusters to capture category-independent variations effectively. Furthermore, CycleGAN~\cite{zhu2017unpaired} uses a cycle-consistent adversarial network, and addresses unpaired training data scenarios, making it possible to perform cross-domain image translation without explicit pixel-wise correspondences.

As diffusion-based methods appears, SphereDiffusion \cite{wu2024spherediffusion} introduces spherical geometry-aware training by enforcing rotational invariance in 360-degree space, ensuring boundary continuity in generated images. 
In contrast, CubeDiff~\cite{kalischek2025cubediff} splits panoramic images into six cube faces, which helps create clearer, distortion-free views compared to traditional equirectangular projections. 
It uses a lightweight solution with special attention and position encoding to generate high-quality panoramic images without complex cross-view attention. It also allows for precise control over details on different cube faces. 
Also, it uses tools (e.g., masks, bounding boxes, etc.) to guide content creation and reconstruct the missing region. 
Furthermore, TwinDiffusion~\cite{zhou2024twindiffusion} enhances coherence and efficiency in panoramic image generation by employing a training-agnostic optimization phase, maximizing similarity between adjacent cropped regions through Crop Fusion and dynamically adjusting sample selection via Cross Sampling. 
This framework establishes a new benchmark in balancing quality and computational efficiency for high-resolution panoramic synthesis. 
These developments highlight a progressive shift from GANs towards diffusion-based architectures, driven by the increasing demand for geometric adaptability, semantic robustness, and high-resolution controllability in panoramic image generation.

However, the above methods primarily focus on generation rather than editing. 
While these generation models have made significant progress in creating high-quality panoramic images, they often lack the capability for precise editing of existing panoramic images. 
\Ours introduces a new paradigm for panoramic image editing, addressing the unique challenges of spherical content and enabling fine-grained modifications by addressing \questionone, \questiontwo, and \questionthree.

\begin{figure}[tb]
    \centering
    \includegraphics[width=\textwidth]{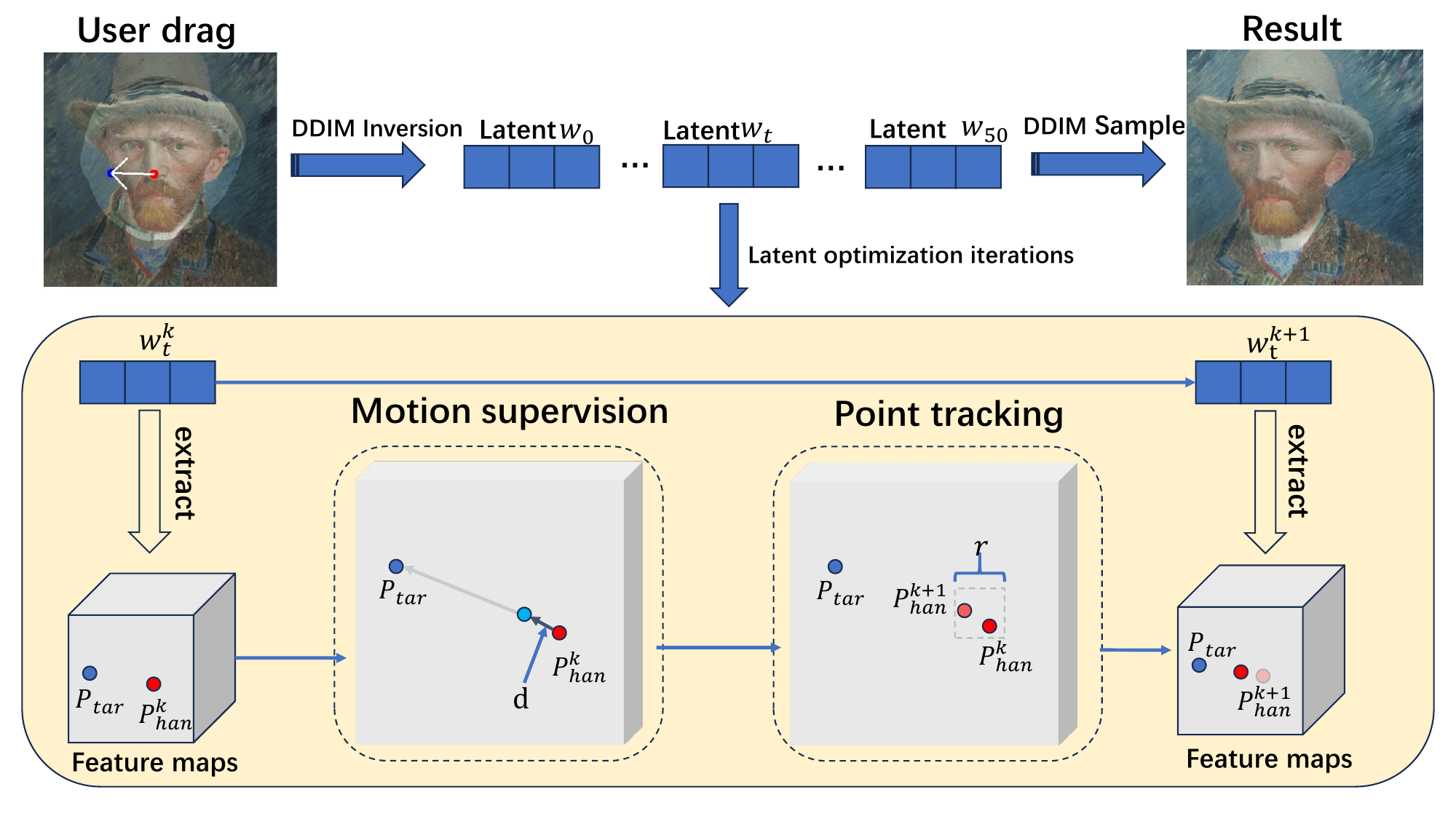}
    \caption{Classic point-interactive image editing pipeline}\label{fig:dragdiff}
\end{figure}

\section{Method}
\label{method}
In this section, we introduce our \Ours in detail.
First, we briefly introduce classic image editing algorithm and build up our notations. 
The notations are summarized in Section A “Notations used in our method” in the Supplementary Material.

Next, we introduce our SphereDrag framework. 
Faced with \questionone challenge, we introduce our \pointone (\pointones) to get the suitable panoramic representation.
Also, we introduce our spherical latent optimization, including \pointtwo (\pointtwos) and \pointthree (\pointthrees), to deal with \questiontwo and \questionthree, respectively.

\subsection{Background and Notations} 
\label{subsec:overview}
As shown in~\cref{fig:dragdiff}, classic point-interactive image editing methods (e.g., DragDiffusion) place a handle point $P_{han}$ (in red), a target point $P_{tar}$ (in blue), and a mask $M$ (denoted by the brighter region of the user drag) on the user drag. 
The $P_{han}$ is the starting point for the motion, while the $P_{tar}$ indicates the desired destination. 
They define the direction and distance of the movement during editing together.
The \(M\) specifies the editable region in the image, while the area outside the mask is the uneditable region.
These methods apply DDIM inversion to obtain a sequence of latent codes, where each code corresponds to the intermediate representation of the image at a specific timestep.
The latent code at fixed timestep $t$, denoted as $w_t$, is used to extract its corresponding UNet feature maps, which encodes the detailed features of the image. 
The feature maps are then iteratively optimized to modify the image according to the editing objective.

During each iteration, the current handle point as \(P_{han}^k\) is moved in the feature maps toward the target point \(P_{tar}\) using \textbf{motion supervision}.
Afterward, the next handle point as \(P_{han}^{k+1}\) is located on the moved feature maps through \textbf{point tracking}.
This process is repeated in each iteration while keeping the region outside the mask unchanged.

\textbf{Motion Supervision}
In motion supervision, we first compute the normalized movement direction from the \(P_{han}^k\) to the \(P_{tar}\) at each iteration $k$. 
The normalized direction ensures that the movement from the handle to the target is consistent and precise, guiding the feature maps to move accordingly in the editing process:
\begin{equation}
    \vec{d} = \frac{P_{tar} - P_{han}^k}{\|P_{tar} - P_{han}^k\|_2},
\end{equation}
where \( \vec{d} \) represents the direction of movement from the current handle point \(P_{han}^k\) to the target point \(P_{tar}\), ensuring a controlled and consistent movement.

Finally, combining the normalized movement direction, the loss function is set as follows:
\begin{align}
    \mathcal{L}_{\text{ms}}(\hat{w}_t^k) =\ &\sum_{q \in \mathcal{F}} \left\| F_{q + \vec{d}}(\hat{w}_t^k) - \text{sg}(F_q(\hat{w}_t^k)) \right\|_1 \nonumber \\
    &+ \lambda \left\| \left(\hat{w}_{t-1}^k - \text{sg}(\hat{w}_{t-1}^0)\right) \odot (1 - M) \right\|_1, 
    \label{eq:motion_supervision}
\end{align}
The motion supervision loss \( \mathcal{L}_{\text{ms}} \) guides the update of the latent code \( \hat{w}_t^k \) during movement.
\textbf{The first term} encourages feature maps extracted at spatial locations \( q \in \mathcal{F} \) to move toward a target position along the motion direction \( \vec{d} \), where \( F_q(\cdot) \) denotes the feature at location \( q = (x, y) \), and \( P_{\text{han}}^k \) is the handle point at the \( k \)-th iteration.
\textbf{The second term} enforces consistency in uneditable regions. 
A binary mask \( M \) indicates editable (\( M=1 \)) and uneditable (\( M=0 \)) areas. 
It penalizes deviations between the current latent code \( \hat{w}_{t-1}^k \) and its initial version \( \hat{w}_{t-1}^0 \) \emph{only in the uneditable regions} (\( M = 0 \)), with the strength controlled by \( \lambda \) and using element-wise multiplication \( \odot \).
The stop-gradient operator \( \text{sg}(\cdot) \) prevents gradients from backpropagating through fixed reference features \( F_q(\hat{w}_t^k) \) and \( \hat{w}_{t-1}^0 \).

\textbf{Point Tracking}
The point tracking process aims to find the next handle point \( P_{han}^{k+1} \) within the search region by minimizing the L1 norm of the difference in features. The search region is defined as:

\begin{equation}
    \Omega_{\text{search}} = \left\{ (x, y) \;\middle|\; |x - p_x| \le r,\ |y - p_y| \le r \right\},\quad p \in \{P_{\text{han}}^k\}.
    \label{eq:feat_region}
\end{equation}
Point tracking is defined as:
\begin{equation}
    P_{han}^{k+1} = \arg \min_{q \in \Omega_{search}} \| F_{q}(\hat{w}^{k+1}_t) - F_{P_{han}}(w^0_t) \|_1,
\end{equation}
where \( F_q(\cdot) \) represents the UNet feature maps extracted from the latent code at position \( q \). 
The next handle point \( P_{han}^{k+1} \) is the point in the region \(\Omega_{search} \) that best matches the feature maps of the handle point \( P_{han} \).
The base radius \( r \) determines the size of the search region used in point tracking.
\( \hat{w}^{k+1}_t \) denotes the latent code at the \( (k+1) \)-th iteration, and \( w^0_t \) denotes the initial latent code at timestep \( t \). 
\( F_{P_{han}}(\cdot) \) represents the feature maps extracted from the latent code at the handle point \( P_{han} \). 
It is a reference for the feature comparison to track the movement of the handle point across iterations.

\begin{figure}[tb]
    \centering
    \includegraphics[width=\textwidth]{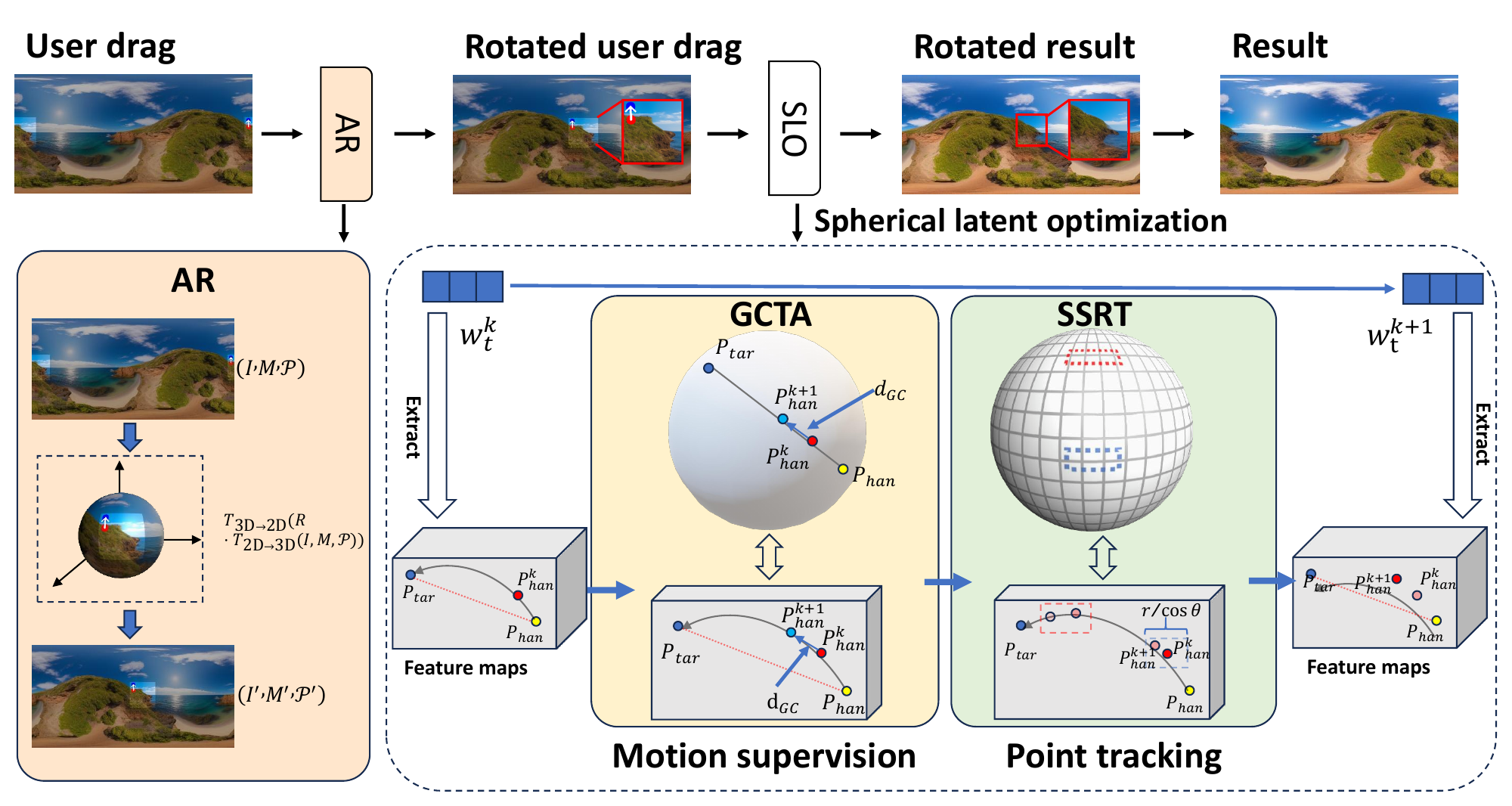}
    \caption{
        Overview of \Ours.  
        Using DragDiffusion as our baseline, we introduce our three parts: \pointone (\pointones), \pointtwo (\pointtwos), and \pointthree (\pointthrees).  
        \textbf{\pointones:} It applies spherical rotation to transform input panoramic images into a suitable representation.  
        \textbf{\pointtwos:} It handles the points \( P_{\text{tar}} \), \( P_k \), and \( P_{\text{han}} \) using the great-circle distance \( d_{\text{gc}} \) in a spherical manner. The underlying planar feature maps visualize the corresponding trajectory.  
        \textbf{\pointthrees:} It highlights the current (blue) and future (red) search regions, which have equal sizes on the sphere. However, their projected areas differ on the feature maps due to spherical distortion.
    }
    \label{fig:spheredrag-pipeline}
\end{figure}

\subsection{SphereDrag Framework}
\label{subsec:spheredrag}
When applied to panoramic images, classic point-interactive image editing faces three major panoramic challenges due to its reliance on planar assumptions: \questionone, \questiontwo, and \questionthree.
As illustrated in~\cref{fig:spheredrag-pipeline}, we introduce three corresponding modules to address these limitations.
\textbf{\pointones} is proposed to deal with \questionone and achieve better image representation. 
\textbf{\pointtwos} and \textbf{\pointthrees} are designed to tackle \questiontwo and \questionthree in spherical latent optimization, respectively.

\subsubsection{Adaptive Reprojection (\pointones)}
\label{subsec:pointone}
As shown in~\cref{fig:fig1}, trajectory discontinuity poses challenges to image editing: trajectories that cross the left and right image boundaries may appear as two disconnected segments.
Inspired by the concept of \emph{spherical symmetry}, the property that any rotation of the original sphere preserves the semantic content of a panoramic image, we introduce a spherical alignment strategy to mitigate these issues.

First, we compute the midpoint \( P_{mid} = (i_m, j_m) \) of the handle point \(P_{han} = (i_1, j_1) \) and the target point \( P_{tar} = (i_2, j_2) \). 
This \(P_{mid}\) is projected from 2D coordinates to spherical coordinates \((\theta, \phi)\) by a spherical projection function \( T_{\text{2D} \rightarrow \text{Sphere}}(\cdot) \).

We then compute the rotation matrix \( R = \mathcal{R}_{\text{align}}(\theta, \phi) \) that aligns the midpoint direction to get a suitable panoramic representation, which is minimally affected by spherical issues (e.g., the center of the panoramic image). 
This rotation is performed on the sphere for the image \( I \), the mask \( M \), and the handle point \( P_{\text{han}} \) and target point \( P_{\text{tar}} \) within \( \mathcal{P} \).
Then, the rotated results are projected back to 2D using \( T_{\text{Sphere} \rightarrow \text{2D}}(\cdot) \). 
The overall spherical alignment process is computed as follows:
\begin{equation}
    (I', M', \mathcal{P}') = \mathcal{T}_{\text{align}}(I, M, \mathcal{P}),
    \label{eq:alignment_pipeline}
\end{equation}
\begin{equation}
    \mathcal{T}_{\text{align}}(I, M, \mathcal{P}) = T_{\text{Sphere} \rightarrow \text{2D}} \left( R \cdot T_{\text{2D} \rightarrow \text{Sphere}}(I, M, \mathcal{P}) \right),
\end{equation}
where the function \( \mathcal{T}_{\text{align}}(\cdot) \) denotes the complete alignment pipeline. 
\( I' \) denotes the image, \( M' \) denotes the mask, and \( \mathcal{P}' \) denotes the handle point and target point. 
$T_{\text{2D} \rightarrow \text{Sphere}}(\cdot)$ is the transformation that projects the input data \( (I, M, \mathcal{P}) \) from 2D coordinates to spherical coordinates.
The rotation matrix \( R \) is applied to these spherical coordinates to align them with a suitable panoramic representation.
Finally, \( T_{\text{Sphere} \rightarrow \text{2D}}(\cdot) \) projects the rotated data back to 2D coordinates, resulting in the final image, mask,handle point and target point, \( (I', M', \mathcal{P}') \).
As shown in~\cref{fig:spheredrag-pipeline}, our spherical alignment strategy makes the editing region more stable semantic consistency and trajectory continuity. 
The further details are provided in Section B “Detail for $\mathcal{T}_{\text{align}}(\cdot)$ in Adaptive Reprojection (AR)” in the Supplementary Material.

\subsubsection{Great-Circle Trajectory Adjustment (\pointtwos)}
\label{subsec:gc}

In classic image editing, motion supervision typically computes the straight movement direction between the current handle point and the target point based on 2D planar assumptions. 
However, when applied to panoramic images, this planar approximation deviates from the correct direction on the sphere, resulting in motion distortions.

To mitigate this challenge, we replace the 2D straight direction with the spherical great-circle direction. 
Specifically, at the $k$-th iteration, given the current handle point $P_{han}^k$, the initial handle point $P_{han}$, and the target point $P_{tar}$, the motion direction is defined as:
\begin{equation}
\vec{d} = \mathcal{G}_{\text{gc}}(P_{han}^k, P_{han}, P_{tar}),
\end{equation}
where $\mathcal{G}_{\text{gc}}(\cdot)$ denotes the great-circle direction computation function (see Section C “Detail for Great Circle Direction Calculation” in the Supplementary Material). 

Based on this direction, the motion supervision loss is defined as:
\begin{align}
    \mathcal{L}_{\text{ms}}(\hat{w}_t^k) =\ &\sum_{q \in 	\mathcal{F}} \left\| F_{q + \vec{d}}(\hat{w}_t^k) - \text{sg}(F_q(\hat{w}_t^k)) \right\|_1 \nonumber \\
    &+ \lambda \left\| \left(\hat{w}_{t-1}^k - \text{sg}(\hat{w}_{t-1}^0)\right) \odot (1 - M) \right\|_1. 
    \label{eq:motion_supervision_gc}
\end{align}
By using the great-circle direction $\vec{d}$, the original planar motion supervision is naturally adapted to the spherical domain. 
This ensures that the movement follows the correct direction on the sphere instead of a straight line on the 2D plane, effectively eliminating the deviation introduced by the planar approximation.
The other variables follow the formulation in~\cref{eq:motion_supervision}.

Compared to conventional planar motion supervision, our \pointtwos more accurately captures directional behavior on the spherical domain, particularly in high-latitude regions where projection distortion is most severe.

\subsubsection{Spherical Search Region Tracking (\pointthrees)}
\label{subsec:point_tracking_la}

In classic image editing, the search region for point tracking is typically defined as a fixed-size square window, based on planar assumptions.
However, when applied to panoramic images, such a design fails to account for the non-uniform distortions introduced by ERP.
Consequently, the search region becomes overstretched in the high-latitude area, resulting in a degraded tracking accuracy, which constitutes the challenge officially defined as \questionthree.

To address this issue, we propose \pointthrees, which adaptively adjusts the search region according to the spherical location of the current handle point. 
Specifically, we project the current handle point \(P_{han}^k\) from 2D planar coordinates to spherical coordinates $(\theta, \phi)$ using the mapping function $T_{\text{2D} \rightarrow \text{Sphere}}(\cdot)$, where $\theta$ and $\phi$ denote longitude and latitude, respectively.

To preserve consistent pixel density across latitudes, we scale the vertical radius of the tracking region as a function of latitude:
\begin{equation}
    r(\phi) = \frac{r}{\cos \phi},
    \label{eq:adaptive_radius}
\end{equation}
where \( r \) is a fixed base radius. The use of \( \cos \phi \) compensates for the horizontal stretching caused by the ERP in high-latitude areas, improving the geometric consistency of the search region and improving the accuracy of the tracking. 
An illustration of how \( \cos \phi \) relates to latitude distortion is provided in Section D “Stretching Factor and Latitude Distortion” in the Supplementary Material.

Based on this, we define the spherical-adaptive search region centered at the current handle point \(P_{han}^k\) as:
\begin{equation}
    \Omega_{\text{search}} = \left\{ (x, y) \ \middle|\ |x - p_x| \le r_0,\ |y - p_y| \le r(\phi) \right\}, \quad p \in P_{han}^k.
    \label{eq:adaptive_region}
\end{equation}
Within this region \(\Omega_{\text{search}}\), the updated handle point is determined by minimizing the L1 distance between UNet features:
\begin{equation}
    P_{han}^{k+1} = \arg\min_{q \in \Omega_{\text{search}}} \left\| F_q(\hat{w}_{t}^{k+1}) - F_{P_{han}}(w_t^0) \right\|_1,
    \label{eq:adaptive_point_tracking}
\end{equation}
where \(F_q(\cdot)\) denotes the UNet feature extracted from the latent code at position \(q\), and \(F_{P_{han}}(w_t^0)\) is the reference feature from the initial handle point in the unoptimized latent code.

By incorporating spherical-aware scaling into the search region, our method better aligns with the geometric properties of panoramic images. This adaptive adjustment enhances the robustness and accuracy of handle point tracking, particularly in high-latitude regions where distortions caused by ERP are more pronounced.

\section{Experiments}
\label{experiments}

In this section, we conduct experiments to evaluate \Ours. 
Firstly, we introduce the benchmark, experimental protocols, and evaluation metrics used in our experiments. 
Secondly, we compare the performance of \Ours with state-of-the-art image editing methods. 
Third, we carry out ablation experiments to clarify the contribution of each module. 
Finally, we provide a detailed discussion of the experimental hyperparameters in Section E “Discussion” in the Supplementary Material.

\subsection{Benchmark, Protocols, and Evaluation Metrics}

\textbf{Benchmark} Classic image editing Dragbench benchmarks utilize planar images, which is not suitable for panoramic image editing evaluation for lack of panoramic characteristics. 
We construct \textbf{\OurBench}, a novel benchmark comprising \imgsize panoramic images. 
It includes 81 panoramic images of different scenes, along with the corresponding masks, handle points, and target points for drag-based editing evaluation. 
These images feature a wide variety of objects and scenarios, including windows, sofas, lamps, mountains, coastlines, and more, providing a diverse and realistic benchmark for evaluating panoramic image editing methods. 

\begin{figure}[tb]
    \centering
    \includegraphics[width=\textwidth]{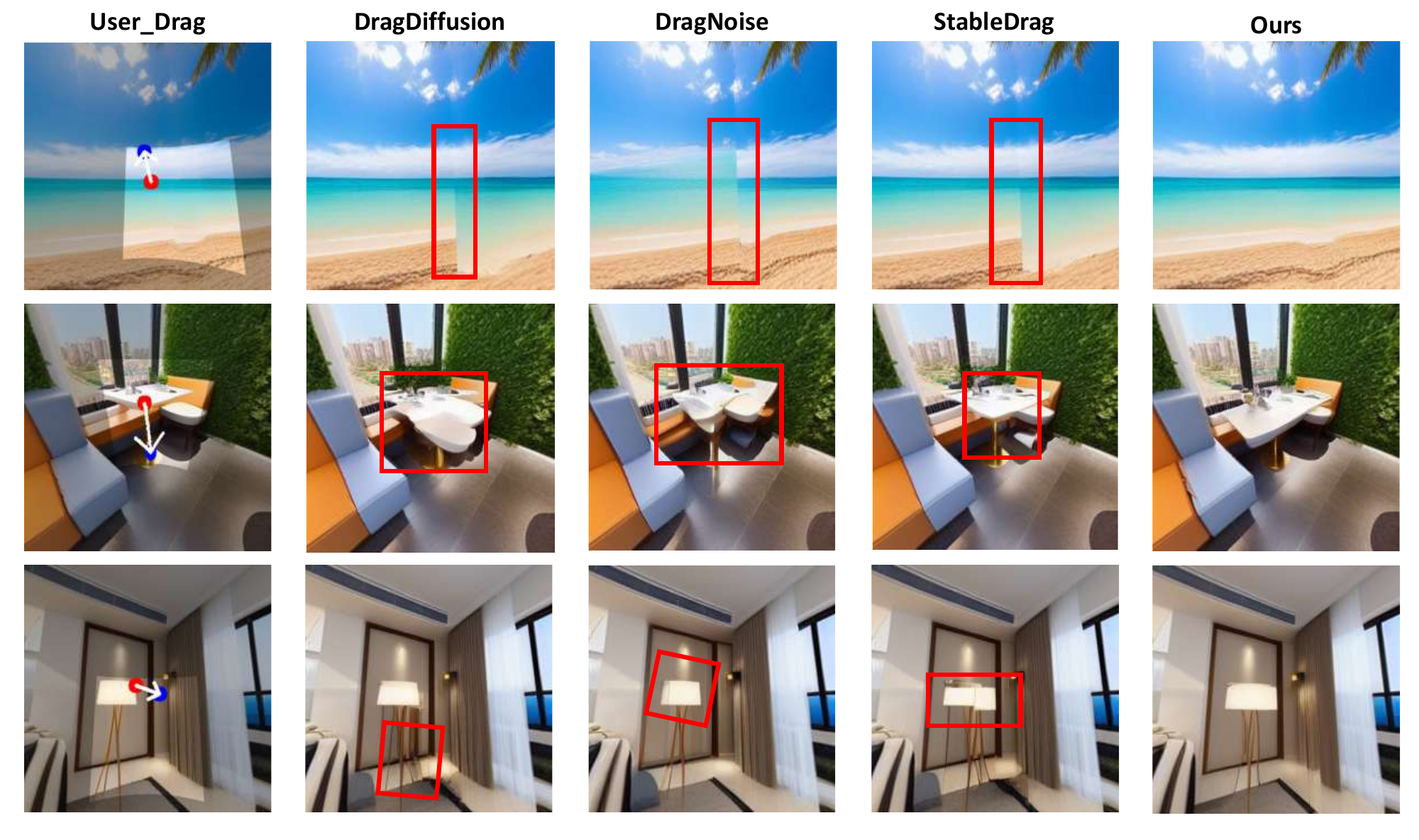}
    \caption{90$^{\circ}$ FOV drag visualization: In the first row, our method effectively handles the seams at the sandy beach with strong winds, while other methods exhibit visible boundary artifacts. 
    In the second row, our method successfully follows the intended dragging path, whereas other methods either fail to complete the drag or produce distorted results. 
    In the third row, our approach accurately interprets the dragging intent. 
    When dragging the lamp, the surrounding elements remain properly arranged, while other methods either fail to move the lamp correctly or result in chaotic outputs. }
    \label{fig:performance}
\end{figure}

\textbf{Evaluation Metrics} Image Fidelity (IF), Fréchet Inception Distance (FID), and Spatial Fréchet Inception Distance (sFID) are used as evaluation metrics. Detailed definitions and formulations of these metrics are provided in Section F “Evaluation Metrics” in the Supplementary Material.

\textbf{Experimental Protocols} All experiments are based on Stable Diffusion v1.5 with a uniform learning rate of 0.01. 
To enhance the model's fidelity for panoramic scenes, the training process involves fine-tuning the cross-attention layers of its UNet architecture via Low-Rank Adaptation (LoRA). 
We set the latent variable at fixed timestep $t$ = 35, the strength of the constraint $\lambda$ = 0.1, the suitable panoramic representation, we set the longitude to 0, The latitude remains consistent before and after rotation and maintain a resolution of the \imgsize image. 
The implementation runs on servers equipped with 8 NVIDIA L40S GPUs.
The rest of the protocols are the same as DragDiffusion.

\subsection{Performance Comparison}
In this section, we evaluate our \Ours method against the latest image editing models, including DragDiffusion~\cite{shi2024dragdiffusion}, DragNoise~\cite{liu2024drag}, and StableDrag~\cite{cui2024stabledrag}.
For this evaluation, we extract perspective views from panoramic images at FOVs of 30$^{\circ}$, 60$^{\circ}$, and 90$^{\circ}$, enabling comprehensive evaluation of model performance across diverse viewing angles.
As shown in ~\cref{tab:quantitative}, at a 30$^{\circ}$ FOV, where the focus is concentrated on the dragged region, the performance differences between the models are particularly pronounced. 
However, as the FOV increases and more undragged areas are incorporated into the generation process, these discrepancies gradually diminish. 
As shown in~\cref{fig:performance}, the figure visualizes the performance variation in different FOVs.

\newcolumntype{C}[1]{>{\centering\arraybackslash}p{#1}}

\begin{table}[tb]
  \centering
  \caption{Quantitative results of our method (\textbf{Ours}) and state-of-the-art models. All methods are evaluated on images with FOV of 30$^{\circ}$, 60$^{\circ}$, and 90$^{\circ}$. Metrics include IF (higher is better), FID, and sFID (lower is better).}
  \label{tab:quantitative}

  \resizebox{\textwidth}{!}{
    \begin{tabular}{l|ccc|ccc|ccc}
    \toprule
    \multirow{2}{*}{Method} 
    & \multicolumn{3}{c|}{30$^{\circ}$} 
    & \multicolumn{3}{c|}{60$^{\circ}$} 
    & \multicolumn{3}{c}{90$^{\circ}$} \\
    & IF$\uparrow$ & FID$\downarrow$ & sFID$\downarrow$
    & IF$\uparrow$ & FID$\downarrow$ & sFID$\downarrow$
    & IF$\uparrow$ & FID$\downarrow$ & sFID$\downarrow$ \\
    \midrule
    Baseline     & 0.6666 & 125.6694 & 409.8609  & 0.8037 & 80.4527  & 277.0527  & 0.8924 & 64.0885 & 182.5420 \\
    DragNoise    & 0.7161 & 126.3004 & 391.0360  & 0.8047 & 88.8964  & 302.2573  & 0.8734 & 77.1288 & 226.8186 \\
    StableDrag   & 0.7133 & 116.4910 & 390.5301  & 0.8312 & 72.2143  & 253.5878  & 0.9064 & 57.7487 & 168.7382 \\
    \textbf{Ours} & \textbf{0.7364} & \textbf{114.8911} & \textbf{375.9909}
                 & \textbf{0.8495} & \textbf{70.4464}   & \textbf{244.3402}
                 & \textbf{0.9191} & \textbf{52.8041}   & \textbf{160.2049} \\
    \bottomrule
  \end{tabular}}
\end{table}

\subsection{Ablation Studies}
In this section, we analyze the contribution of each module to overall performance on images with FOV angles of 30$^{\circ}$, 60$^{\circ}$, and 90$^{\circ}$. 
The evaluation is conducted using the IF metric.
To provide a more thorough assessment, we also incorporate the sFID metric in an extended experiment.

The results are summarized in \cref{tab:ablation-fza}. 
We consider several configurations, including a baseline model (without any our module), a model with \pointone (\pointones), and extended versions incorporating \pointtwo (\pointtwos) and \pointthree (\pointthrees).

The results show that \pointones consistently improves performance across all FOV settings. 
Adding \pointtwos brings further gains, and combining \pointones, \pointtwos, and \pointthrees achieves the best results. 
Section G, “Effectiveness of Adaptive Reprojection,” in the Supplementary Material provides visualizations demonstrating that \pointones significantly enhances semantic continuity and geometric structure in the output panoramic images.

\newcolumntype{Y}{>{\centering\arraybackslash}X}

\begin{table}[tb]
  \centering
  \caption{Ablation study results comparing different model variants on images with FOV of 30$^{\circ}$, 60$^{\circ}$, and 90$^{\circ}$. AR / GCTA / SSRT denote adaptive reprojection, great-circle trajectory adjustment, and spherical search region tracking, respectively.}
  \label{tab:ablation-fza}

  \setlength{\tabcolsep}{7pt}
  
\begin{tabularx}{\textwidth}{YYY|Y|Y|Y|Y|Y|Y}
    \toprule
    \multirow{2}{*}{AR} & \multirow{2}{*}{GCTA} & \multirow{2}{*}{SSRT} 
    & \multicolumn{3}{c|}{IF$\uparrow$} & \multicolumn{3}{c}{sFID$\downarrow$} \\
    \cmidrule{4-9}
    & & & 30$^{\circ}$ & 60$^{\circ}$ & 90$^{\circ}$ & 30$^{\circ}$ & 60$^{\circ}$ & 90$^{\circ}$ \\
    \midrule
    \ding{55} & \ding{55} & \ding{55} & 0.6666 & 0.8037 & 0.8924 & 409.86 & 277.05 & 182.54 \\
    \ding{51} & \ding{55} & \ding{55} & 0.7060 & 0.8321 & 0.9101 & 398.08 & 260.21 & 172.47 \\
    \ding{51} & \ding{51} & \ding{55} & 0.7081 & 0.8326 & 0.9103 & 399.30 & 259.97 & 172.21 \\
    \ding{51} & \ding{51} & \ding{51} & \textbf{0.7364} & \textbf{0.8495} & \textbf{0.9191} & \textbf{375.99} & \textbf{244.34} & \textbf{160.20} \\
    \bottomrule
\end{tabularx}
\end{table}

\section{Conclusion}
Faced with three major challenges in spherical panoramic image editing: \questionone, \questiontwo, and \questionthree.
We propose \Ours, a panoramic image editing framework designed to produce high-quality, easily controllable, and precisely manipulated spherical panoramic images.
For \questionone, we introduce \pointone, placing the region that needs to be manipulated in the suitable location of the panoramic image.
For \questiontwo and \questionthree, we propose our spherical latent optimization composed of \pointtwo and \pointthree to accurately track the movement trajectory of the editing point.
Also, we construct a panoramic image editing benchmark, \OurBench, to compare \Ours with other methods in spherical panoramic image editing task comprehensively. 
Experimental results show that \Ours achieves the state-of-the-art quantitative and qualitative performance.

\section*{Acknowledgements}
This work is supported in part by Natural Science Foundation of Shanghai under Grant 24ZR1425600, 
“Pioneer” and “Leading Goose” R\&D Program of Zhejiang (No.2025C02014), 
Ningbo Science and Technology Special Projects under Grant No. 2025Z028, 
the Chenguang Program of Shanghai Education Development Foundation and Shanghai Municipal Education Commission with Grant No. 24CGA73, 
and the Fundamental Research Funds for the Central Universities.

\newpage

\renewcommand\thesection{\Alph{section}}
\setcounter{section}{0} 
\renewcommand{\theequation}{S-\arabic{equation}}
\setcounter{equation}{0}
\renewcommand{\thefigure}{S-\arabic{figure}}
\setcounter{figure}{0}
\renewcommand{\thetable}{S-\arabic{table}}
\setcounter{table}{0}
\begin{center}
    \Large\bfseries Supplementary Material: ``SphereDrag: Spherical Geometry-Aware Panoramic Image Editing''
\end{center}

\renewcommand\thesection{\Alph{section}}
\setcounter{section}{0} 
\renewcommand{\theequation}{S-\arabic{equation}}
\setcounter{equation}{0}
\renewcommand{\thefigure}{S-\arabic{figure}}
\setcounter{figure}{0}
\renewcommand{\thetable}{S-\arabic{table}}
\setcounter{table}{0}

\section{Notations used in our method}
\begin{table}[htb]
    \centering

    \label{tab:notations}
    \renewcommand\arraystretch{1.2}
    {\large 
    \begin{tabularx}{\textwidth}{@{}lX@{}}
    \toprule
    \( sg(\cdot) \) & Stop gradient operation, preventing gradient backpropagation. \\
    \( P_{\text{han}} / P_{\text{tar}} \) & The handle point / the target point. \\
    \( P_{\text{mid}} \) & The midpoint between the handle and target point. \\
    \( P_{\text{han}}^k \) & The current handle point at the \(k\)-th iteration. \\
    \( r \) & The base radius determining the search region size. \\
    \( (\theta, \phi) \) & The spherical coordinates: longitude \(\theta\), latitude \(\phi\). \\
    \( \mathcal{F} \) & The feature maps. \\
    \( \Omega_{\text{search}} \) & The search region. \\
    \( \hat{w}_t^k \) & The latent variable at fixed timestep \( t \) after the \(k\)-th iteration. \\
    \( w_t^0 \) & The initial latent variable at timestep \( t \). \\
    \( T_{\text{Sphere} \rightarrow \text{2D}}(\cdot) \) & The projection from spherical to 2D coordinates. \\
    \( T_{\text{2D} \rightarrow \text{Sphere}}(\cdot) \) & The projection from 2D to spherical coordinates. \\
    \( M \) & A binary mask indicating editable vs.\ uneditable regions. \\
    \( \vec{d} \) & The direction vector from the current handle point to the target point. \\
    \( \mathcal{L}_{\text{ms}} \) & The motion supervision loss. \\
    \( F_q(\cdot) \) & Feature maps extracted at location \( q \). \\
    \( F_q(\hat{w}_t^k) \) & Feature maps extracted from \( \hat{w}_t^k \) at \( q \), with spherical alignment and resizing. \\
    \( \mathcal{R}_{\text{align}}(\theta, \phi) \) & Rotation to align \((\theta, \phi)\) with panoramic view. \\
    \( \mathcal{G}_{\text{gc}}(\cdot) \) & Great-circle direction computation. \\
    \bottomrule
    \end{tabularx}}
\end{table}

\section{Detail for $\mathcal{T}_{\text{align}}(\cdot)$ in Adaptive Reprojection (AR)}
\label{app1}
In this section, we describe the details of the $\mathcal{T}_{\text{align}}(\cdot)$ process.
Given an input panoramic image with width $W$, height $H$, corresponding handle point $P_{\text{han}}$, target point $P_{\text{tar}}$, and masks $M$.
The handle point in the image is $P_{\text{han}} = (i_1, j_1)$, and the target point is $P_{\text{tar}} = (i_2, j_2)$.
The midpoint $P_{\text{mid}} = (i_m, j_m)$ is the average of the handle and target pixel coordinates $i_m = \frac{i_1 + i_2}{2}, \quad j_m = \frac{j_1 + j_2}{2}$.
Then we converted to spherical coordinates (latitude $\theta$ and longitude $\phi$ as follow:
\begin{equation}
    \theta = \frac{\pi}{2} - \frac{j}{H} \cdot \pi, \quad \phi = \frac{i}{W} \cdot 2 \pi - \pi. 
\end{equation}
 
After calculating the spherical coordinates, we align the image by adjusting the longitude $\phi$ and latitude $\theta$ of the midpoint to match a suitable panoramic representation.

\textbf{Rotation to align the longitude:}
As before, the rotation is used to align the image's midpoint to the desired longitude. This rotation matrix $R_{\phi}$ is given by:
\begin{equation}
    R_{\phi}(\Delta \phi) = \begin{bmatrix}
    \cos(\Delta \phi) & -\sin(\Delta \phi) & 0 \\
    \sin(\Delta \phi) & \cos(\Delta \phi) & 0 \\
    0 & 0 & 1
    \end{bmatrix},
\end{equation}
where $\Delta \phi$ is the angle required to adjust the longitude of the midpoint to the desired longitude.

\textbf{Rotation to align the latitude:}
Once the longitude is aligned, the rotation is used to align the image's midpoint to the desired latitude. This rotation matrix $R_{\theta}$ is given by:
\begin{equation}
    R_{\theta}(\Delta \theta) = \begin{bmatrix}
    1 & 0 & 0 \\
    0 & \cos(\Delta \theta) & -\sin(\Delta \theta) \\
    0 & \sin(\Delta \theta) & \cos(\Delta \theta)
    \end{bmatrix},
\end{equation}
where $\Delta \theta$ is the angle required to adjust the latitude of the midpoint to the desired latitude. 

\textbf{Combined Rotation Matrix:} 
The final rotation matrix $R$ is the combination of both rotations. The combined rotation matrix is given by the matrix multiplication:
\begin{equation}
    R(\Delta \theta, \Delta \phi) = R_{\theta}(\Delta \theta) \cdot R_{\phi}(\Delta \phi),
\end{equation}

Finally, the rotation matrix \( R \) is applied to each pixel \( i \) in the panoramic image \( I \), the mask \( M \), and the point set \( \mathcal{P} \), which includes the handle and target points. Since the mask and points are defined in the same coordinate system as the image, applying \( R \) to these elements preserves their spatial relationships. The 3D vector \( \mathbf{p}_i = (x_i, y_i, z_i) \) for each pixel is rotated to obtain the new 3D vector \( \mathbf{p}_i^{\text{new}} = R \cdot \mathbf{p}_i \). These new 3D coordinates are then projected back onto the image plane using inverse spherical mapping, resulting in the transformed image \( I' \), mask \( M' \), and points \( \mathcal{P}' \).

\section{Detail for Great Circle Direction Calculation}
\label{app2}
In this section, we describe the great-circle movement direction computation in detail. 
This method approximates the optimal direction for moving from a current position toward a target along a great circle on a sphere.

Given the handle point $P_{\text{han}}$, the target point $P_{\text{tar}}$, and the current handle point $P_{\text{han}}^k$ in panoramic image 2D coordinates, we first convert these points into spherical coordinates, and then into 3D coordinates to facilitate the great-circle calculations.

The conversion from panoramic image coordinates to spherical coordinates (latitude $\theta$ and longitude $\phi$) for a pixel at position $(i, j)$ in panoramic image (width $W$ and height $H$) coordinates is given by:
\begin{equation}
    \theta = \frac{\pi}{2} - \frac{j}{H} \cdot \pi, \quad \phi = \frac{i}{W} \cdot 2\pi - \pi,
    \label{appeq1}
\end{equation}
where $\theta \in [-\frac{\pi}{2}, \frac{\pi}{2}]$, $\phi \in (-\pi, \pi]$, $\theta = 0$ indicating equatorial, and $\phi= 0$ indicating meridian (the middle of the panoramic image). 

These spherical coordinates are then transformed into 3D Cartesian coordinates on the unit sphere using the standard conversion:
\begin{equation}
    \begin{aligned}
        x &= \cos(\theta) \cos(\phi), \\
        y &= \cos(\theta) \sin(\phi), \\
        z &= \sin(\theta).
    \end{aligned}
    \label{appeq2}
\end{equation}

This yields the 3D unit vectors $\mathbf{P}_{\text{han}}$, $\mathbf{P}_{\text{tar}}$, and $\mathbf{P}_{\text{han}}^k$ corresponding to the handle point, target point, and current handle point, respectively.

To compute the great-circle movement direction between the handle and target points, we first determine the unit normal vector to the plane containing the great circle. 
This unit normal vector $\mathbf{n}$ is given by the normalized cross product of the handle and target vectors:
\begin{equation}
    \mathbf{n} = \frac{\mathbf{P}_{\text{han}} \times \mathbf{P}_{\text{tar}}}{\|\mathbf{P}_{\text{han}} \times \mathbf{P}_{\text{tar}}\|}.
\end{equation}

Since the current position $\mathbf{P}_{\text{han}}^k$ may not lie exactly on this great circle, we orthogonally project it onto the plane defined by $\mathbf{n}$:
\begin{equation}
    \mathbf{P}' = \frac{\mathbf{P}_{\text{han}}^k - (\mathbf{P}_{\text{han}}^k \cdot \mathbf{n}) \mathbf{n}}{\|\mathbf{P}_{\text{han}}^k - (\mathbf{P}_{\text{han}}^k \cdot \mathbf{n}) \mathbf{n}\|}.
\end{equation}
This projection $\mathbf{P}'$ lies on the unit sphere and is the closest point on the great-circle path to the current position.

The desired movement direction in 3D space is then computed as the normalized vector from $\mathbf{P}'$ to the target:
\begin{equation}
\mathbf{v}_{\text{move}} = \frac{\mathbf{P}_{\text{tar}} - \mathbf{P}'}{\|\mathbf{P}_{\text{tar}} - \mathbf{P}'\|}.
\end{equation}

To interpret this 3D direction in a local 2D coordinates, we construct an orthonormal basis on the tangent plane of the sphere at the current position $\mathbf{P}_{\text{han}}^k$. 
This basis consists of two vectors, the zonal (eastward) vector $\hat{\mathbf{e}}_\phi$ and the meridional (northward) vector $\hat{\mathbf{e}}_\theta$, their unit vectors are: 

\begin{align}
    \hat{\mathbf{e}}_\phi &= \frac{1}{\cos(\theta)} \frac{\partial \mathbf{P}}{\partial \phi} = \left[
                                    -\sin(\phi),\
                                    \cos(\phi),\
                                    0
                                 \right]^T, \\
    \hat{\mathbf{e}}_\theta &= \frac{\partial \mathbf{P}}{\partial \theta} = \left[
            -\sin(\theta)\cos(\phi),\
            -\sin(\theta)\sin(\phi),\
            \cos(\theta)
           \right]^T.
\end{align}

Project $\mathbf{v}_{\text{move}}$ onto this tangent plane and make vector decomposition based on the orthonormal basis above:
\begin{equation}
    \begin{bmatrix}
        v_\phi \\
        v_\theta

    \end{bmatrix}
    =
    \begin{bmatrix}
        \mathbf{v}_{\text{move}} \cdot \hat{\mathbf{e}}_\phi \\
        \mathbf{v}_{\text{move}} \cdot \hat{\mathbf{e}}_\theta
        
    \end{bmatrix}.
\end{equation}

the tangent plane at $\mathbf{P}_{\text{han}}^k$ does not correspond to the panoramic image plane, but the zonal (eastward) component are scaled by $1 / \cos(\theta)$ due to the ERP: 

\begin{equation}
    \vec{d} = \frac{[\frac{v_\phi}{\cos(\theta)}, v_\theta]}{\left\|\left[\frac{v_\phi}{\cos(\theta)}, v_\theta \right]\right\|}
                                         .
\end{equation}

This normalized vector $\vec{d}$ represents the final movement direction in the panoramic image coordinate and can be directly interpreted as a direction vector in pixel space.

\section{Stretching Factor and Latitude Distortion}
\begin{figure}[H]
    \centering
    \includegraphics[width=\textwidth]{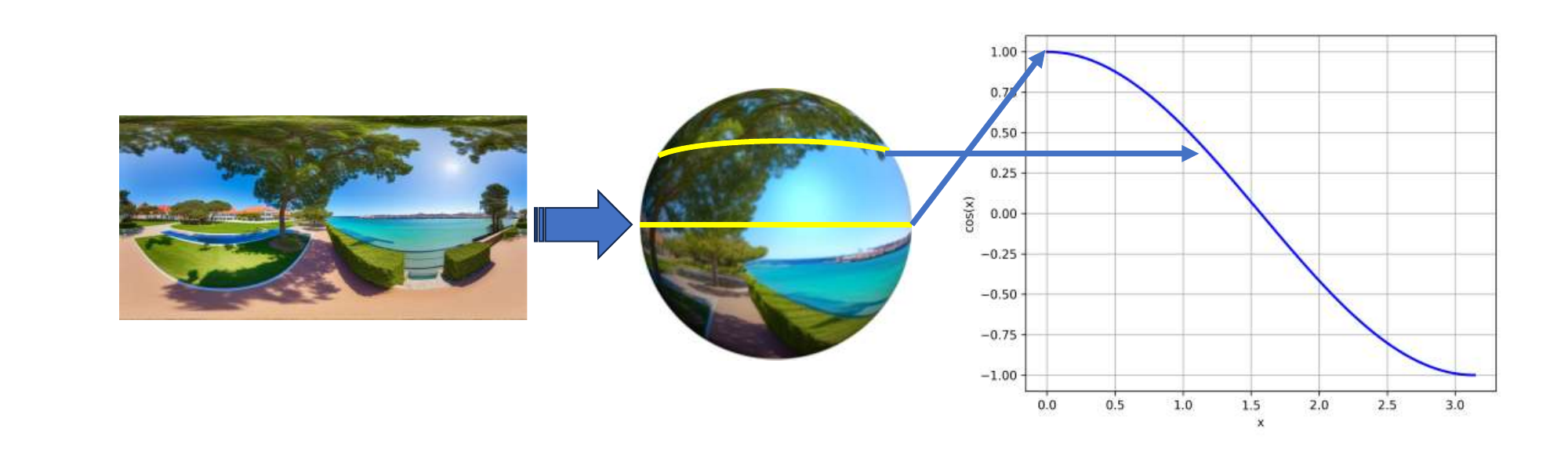}
    \caption{In a panoramic image, \(\cos \phi\) acts as the stretching factor, ensuring that the degree of stretching in the latitude direction from the sphere to the plane is consistent with the spherical geometry. The variation of \(\cos \phi\) with latitude \(\phi\) precisely reflects the projection requirements of the parallel lengths on the sphere, thus maintaining the accuracy and consistency of geographic information during the projection process.}
    \label{fig:figcos}
\end{figure}

\section{Evaluation Metrics}
\begin{itemize}
    \item \textbf{IF} is a metric that quantifies the visual similarity between the original and edited images. It is calculated by first computing the LPIPS~\cite{zhang2018unreasonable} values between the original and edited images, averaging these values, and then subtracting the average LPIPS score from 1:
    \begin{equation}
        \text{IF} = 1 - \text{avg (LPIPS)},
    \end{equation}
    where \(\text{avg(LPIPS)}\) denotes the mean LPIPS value over all image patches or samples. A higher IF value indicates better preservation of the original image's identity.

    \item \textbf{FID}~\cite{heusel2017gans} is a widely used metric for evaluating generative models. 
    It measures the similarity between the feature distributions of real and generated samples in a feature space. 
    The calculation involves extracting features using the Inception-v3 model, computing the mean ($\mu$) and covariance matrix ($\Sigma$) for both real and generated samples, and then computing FID as:
    \begin{equation}
        \text{FID} = \|\mu_1 - \mu_2\|_2^2 + \text{Tr}\left(\Sigma_1 + \Sigma_2 - 2 \left(\Sigma_1 \Sigma_2\right)^{1/2}\right),
    \end{equation}
    where $\mu_1, \Sigma_1$ are the mean and covariance of real samples, and $\mu_2, \Sigma_2$ are those of generated samples.

    \item \textbf{sFID}~\cite{nash2021generating} is a variant of FID that uses spatial features instead of global features. 
    Like FID, it compares the feature distributions of real and generated samples, but focuses on spatial information rather than pooled features. 
    The computation formula is identical to FID, applied over spatial features.
\end{itemize}

IF indicates editing accuracy when both FID and sFID show the structure and semantic preservation capabilities of models. 
The multimetric evaluation considers both editing quality and panoramic structure maintenance.

\begin{figure}[H]
    \centering
    \includegraphics[width=\textwidth]{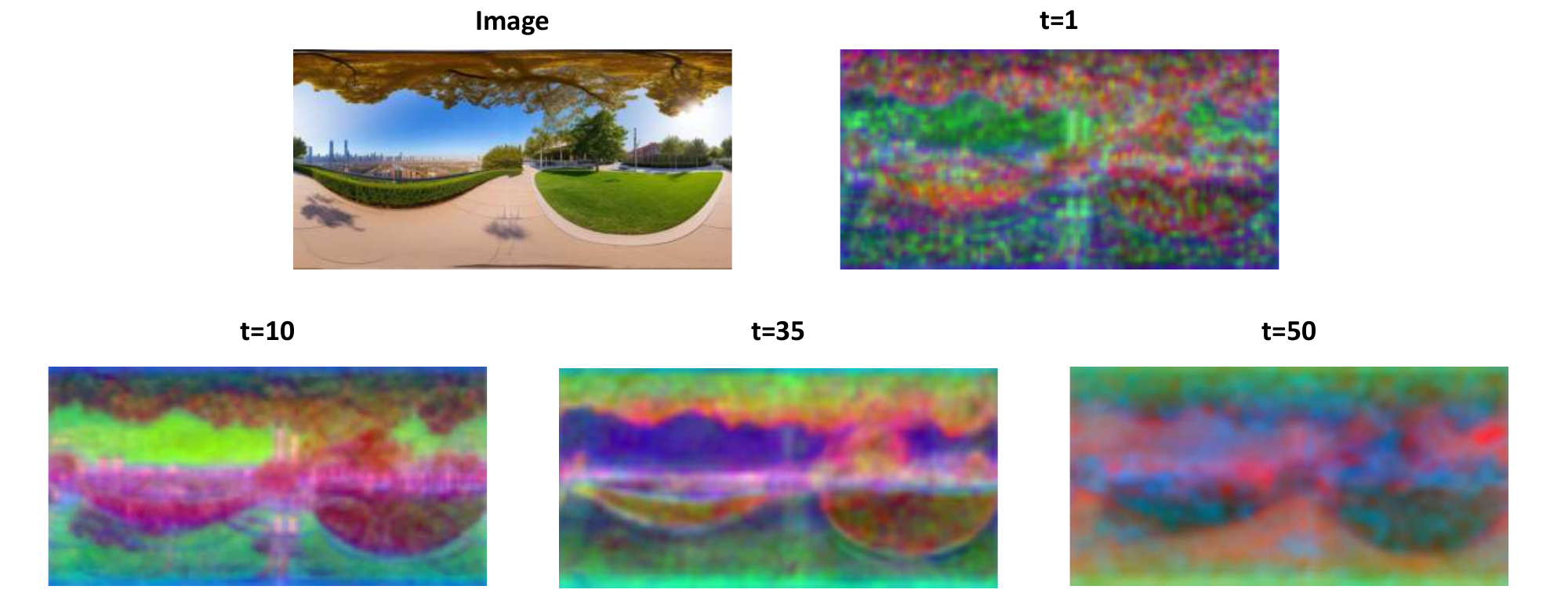}
    \caption{PCA visualization results. At smaller timesteps (e.g., $t=1, 10$), the latent variable contains contains redundant details, while at larger timesteps (e.g., $t=50$), the spatial information is overly diminished. 
    At $t=35$, a good balance is achieved with fewer noise and richer spatial information, effectively preserving  spatial coherence.}
    \label{fig:pca}
\end{figure}

\begin{figure*}[htb]
    \centering
    \begin{tabular}{ccc}
        \begin{subfigure}{0.3\linewidth}
            \includegraphics[width=\linewidth]{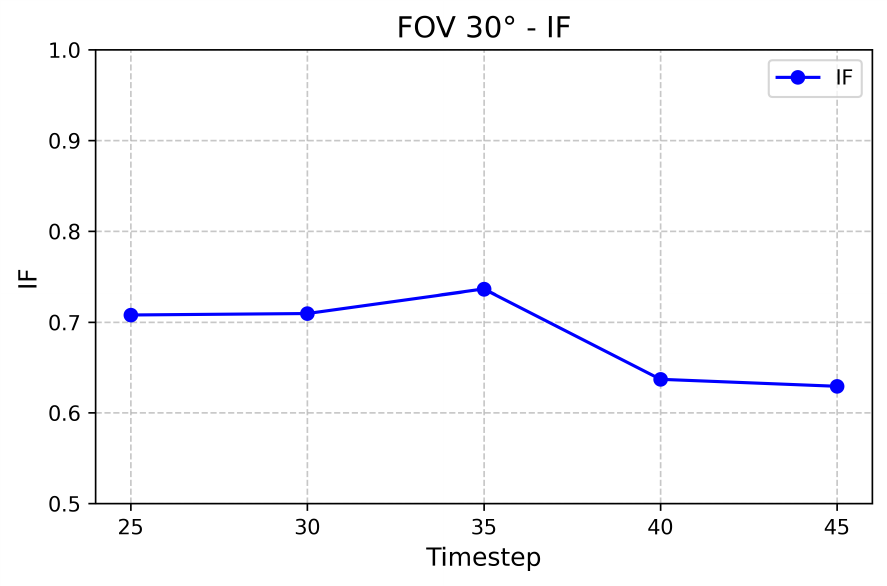}
            \caption{FOV 30°: $t$ vs. IF}
            \label{fig:timestep:a}
        \end{subfigure} &
        \begin{subfigure}{0.3\linewidth}
            \includegraphics[width=\linewidth]{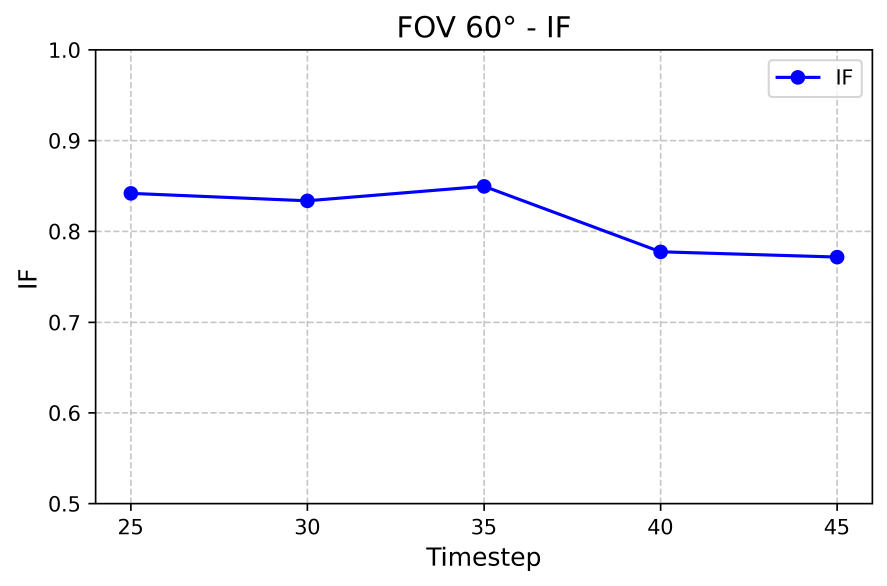}
            \caption{FOV 60°: $t$ vs. IF}
            \label{fig:timestep:b}
        \end{subfigure} &
        \begin{subfigure}{0.3\linewidth}
            \includegraphics[width=\linewidth]{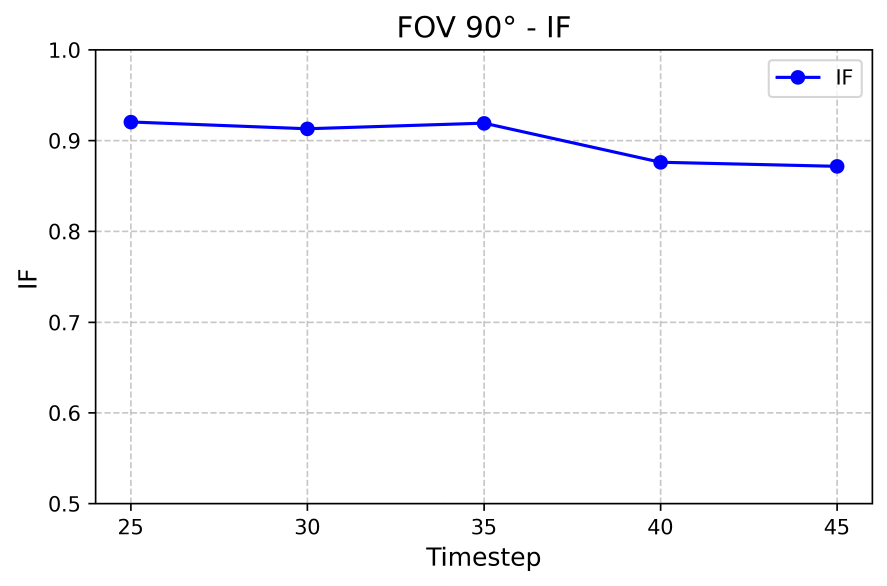}
            \caption{FOV 90°: $t$ vs. IF}
            \label{fig:timestep:c}
        \end{subfigure} \\
        \begin{subfigure}{0.3\linewidth}
            \includegraphics[width=\linewidth]{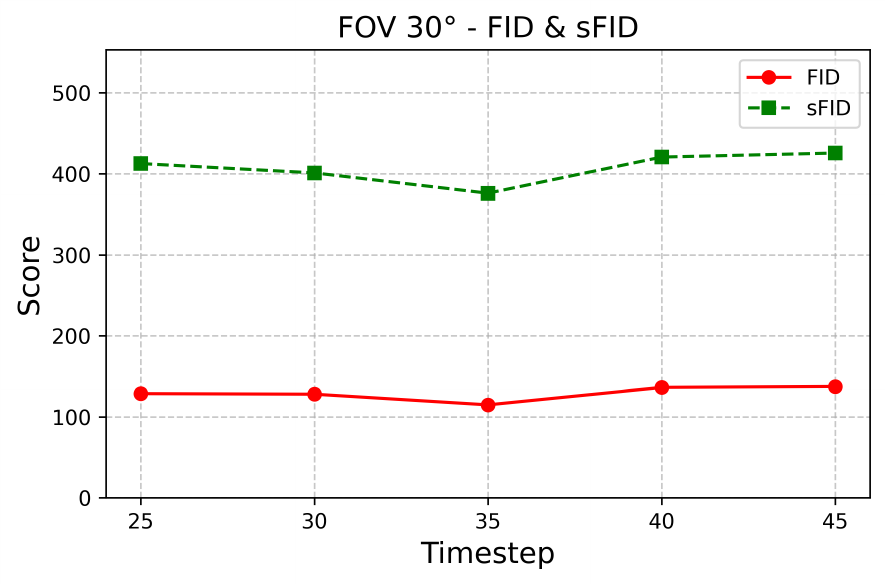}
            \caption{FOV 30°: $t$ vs. FID \& sFID}
            \label{fig:timestep:d}
        \end{subfigure} &
        \begin{subfigure}{0.3\linewidth}
            \includegraphics[width=\linewidth]{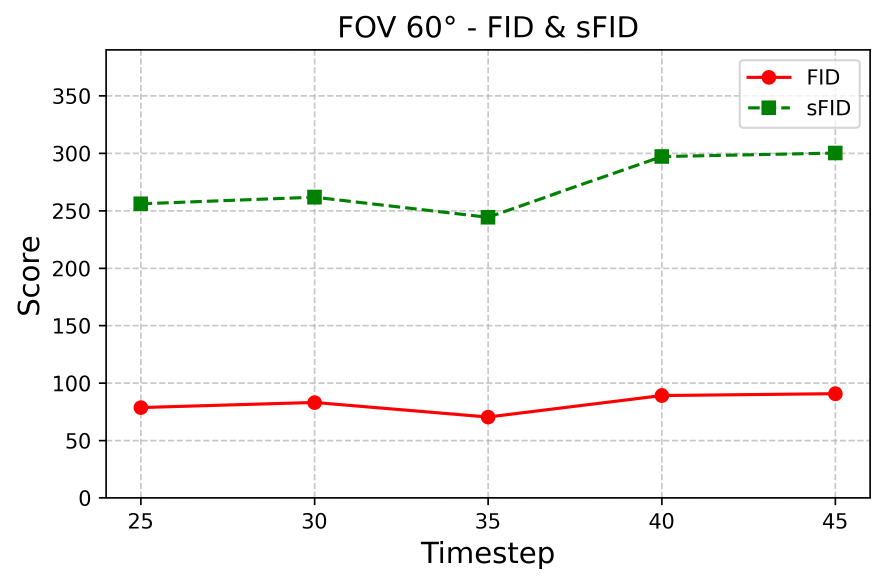}
            \caption{FOV 60°: $t$ vs. FID \& sFID}
            \label{fig:timestep:e}
        \end{subfigure} &
        \begin{subfigure}{0.3\linewidth}
            \includegraphics[width=\linewidth]{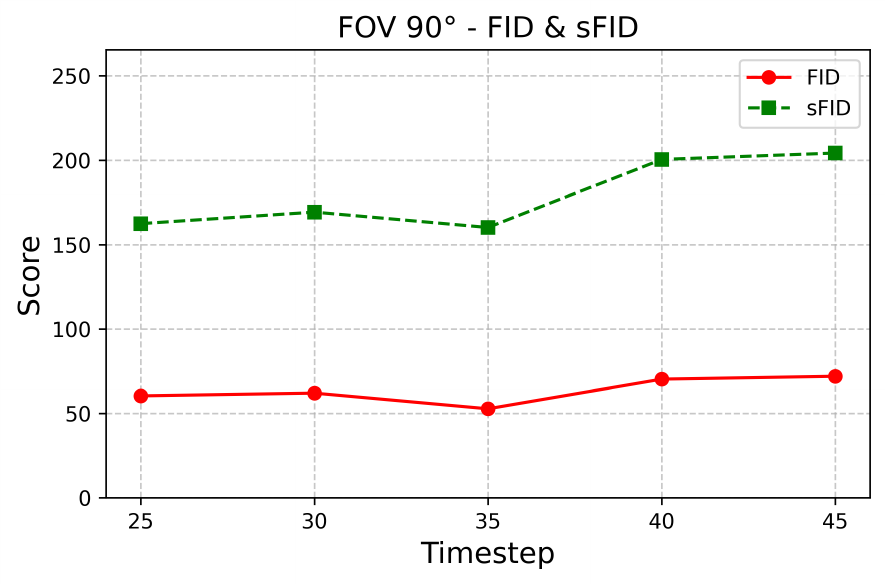}
            \caption{FOV 90°: $t$ vs. FID \& sFID}
            \label{fig:timestep:f}
        \end{subfigure} \\
    \end{tabular}
    \caption{Influence of different hyperparameters $t$ across various FOV settings (30$^{\circ}$, 60$^{\circ}$, and 90$^{\circ}$). The top row (a-c) shows the relationship between $\lambda$ and the IF value, while the bottom row (d-f) shows $\lambda$ versus FID and sFID for the corresponding FOV.}
    \label{fig:timestep}
\end{figure*}

\begin{figure*}[htb]
    \centering
    \begin{tabular}{ccc}
        \begin{subfigure}{0.3\linewidth}
            \includegraphics[width=\linewidth]{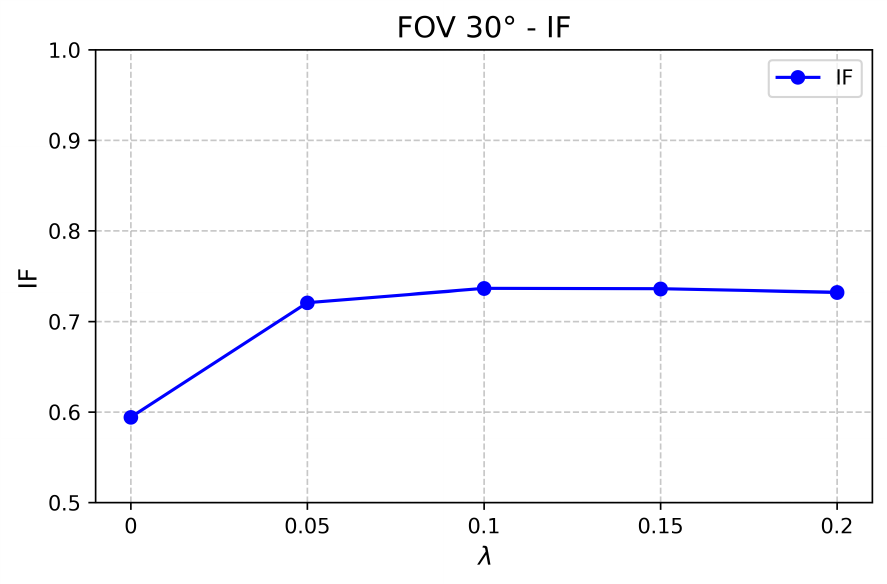}
            \caption{FOV 30°: $\lambda$ vs. IF}
            \label{fig:lam:a}
        \end{subfigure} &
        \begin{subfigure}{0.3\linewidth}
            \includegraphics[width=\linewidth]{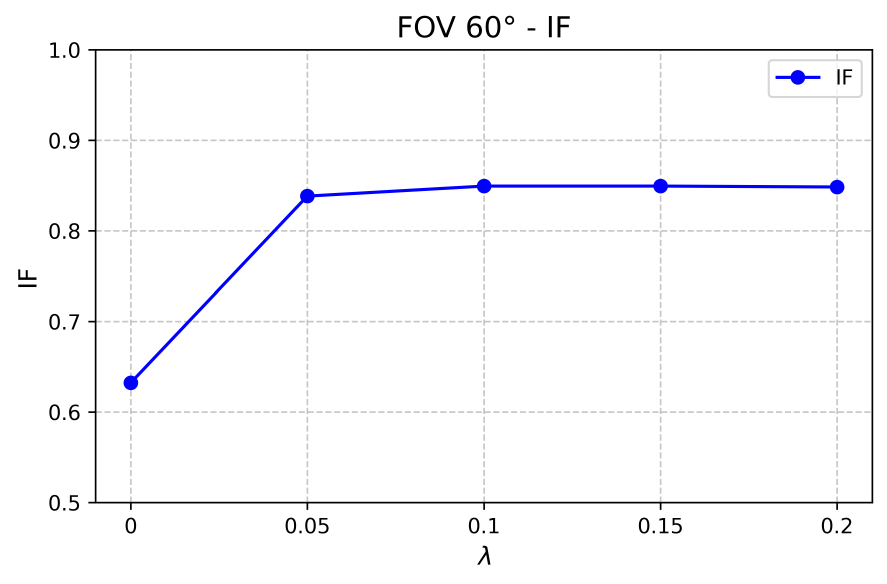}
            \caption{FOV 60°: $\lambda$ vs. IF}
            \label{fig:lam:b}
        \end{subfigure} &
        \begin{subfigure}{0.3\linewidth}
            \includegraphics[width=\linewidth]{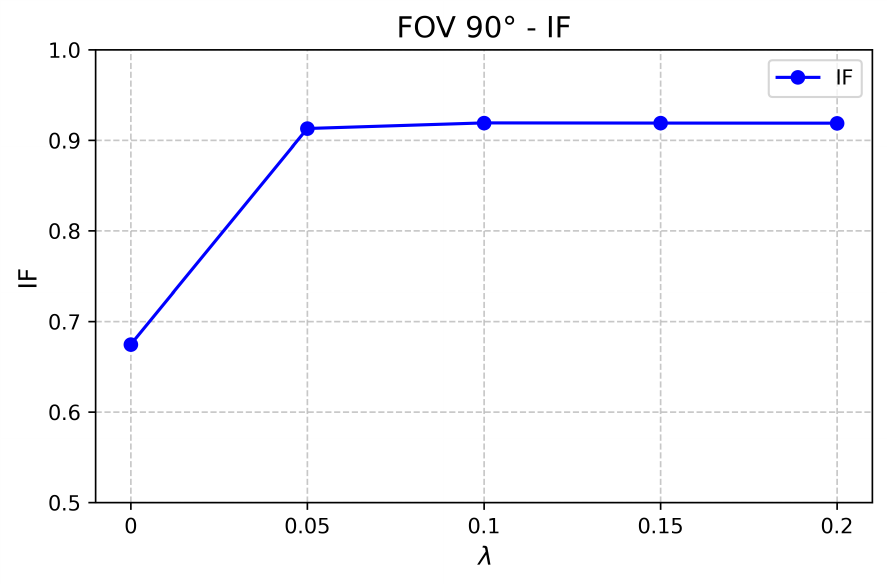}
            \caption{FOV 90°: $\lambda$ vs. IF}
            \label{fig:lam:c}
        \end{subfigure} \\
        \begin{subfigure}{0.3\linewidth}
            \includegraphics[width=\linewidth]{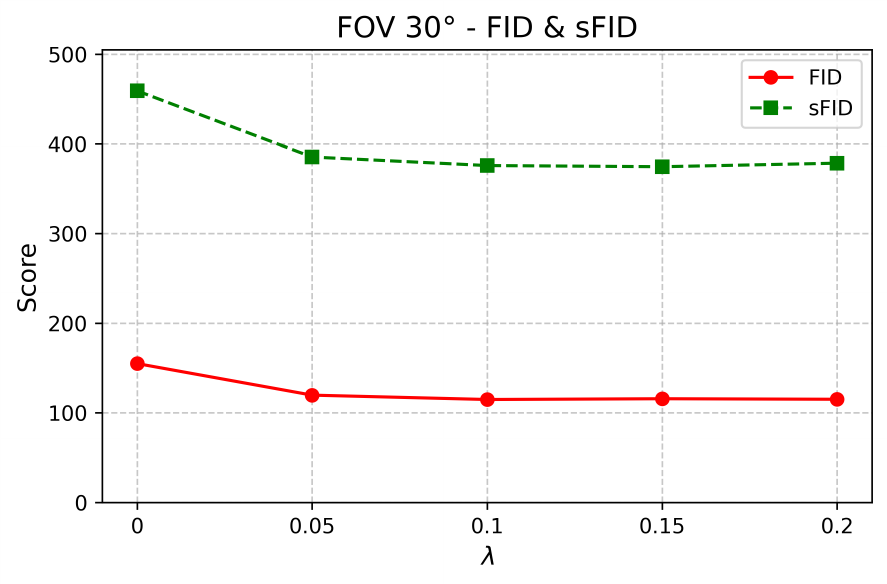}
            \caption{FOV 30°: $\lambda$ vs. FID \& sFID}
            \label{fig:lam:d}
        \end{subfigure} &
        \begin{subfigure}{0.3\linewidth}
            \includegraphics[width=\linewidth]{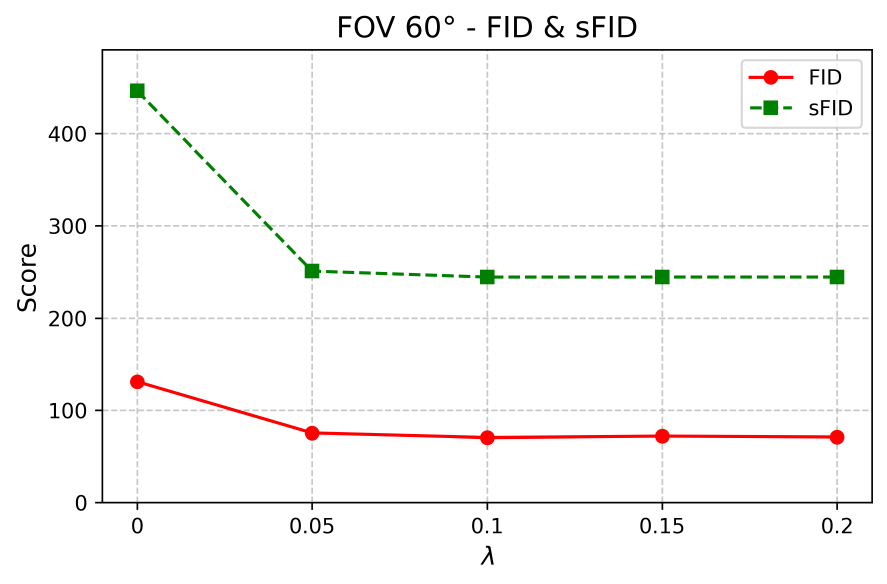}
            \caption{FOV 60°: $\lambda$ vs. FID \& sFID}
            \label{fig:lam:e}
        \end{subfigure} &
        \begin{subfigure}{0.3\linewidth}
            \includegraphics[width=\linewidth]{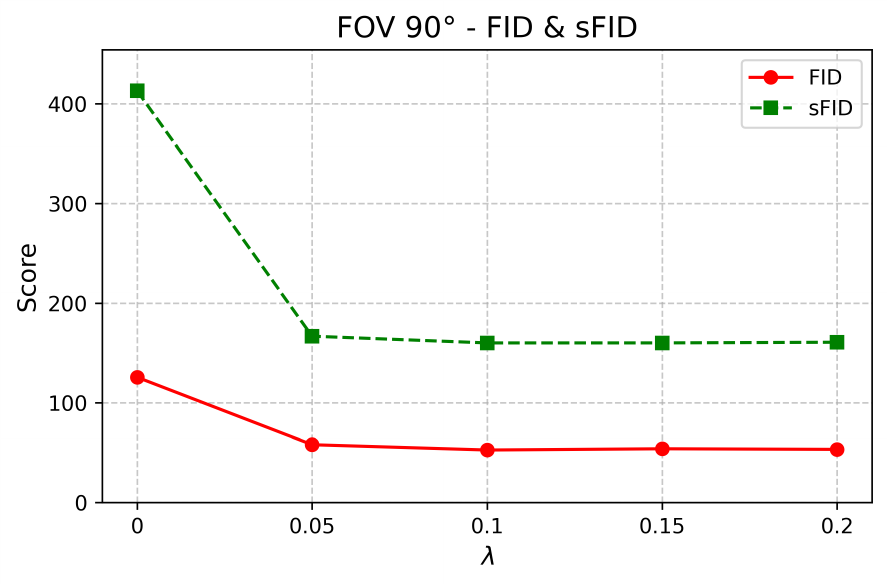}
            \caption{FOV 90°: $\lambda$ vs. FID \& sFID}
            \label{fig:lam:f}
        \end{subfigure} \\
    \end{tabular}
    \caption{Influence of different hyperparameters $\lambda$ across various FOV settings (30$^{\circ}$, 60$^{\circ}$, and 90$^{\circ}$). The top row (a-c) shows the relationship between $\lambda$ and the IF value, while the bottom row (d-f) shows $\lambda$ versus FID and sFID for the corresponding FOV.}
    \label{fig:lam}
\end{figure*}
\section{Discussion}

In this section, we discuss the influence of different hyperparameter $t$ and the $\lambda$. 

\textbf{Hyperparameters $t$}
As shown in~\cref{fig:timestep}, based on the evaluation in three FOVs (30$^{\circ}$, 60$^{\circ}$ and 90$^{\circ}$), $ t=35 $ demonstrates consistently strong performance in IF, FID and sFID, with noticeable improvements in narrower FOVs (30$^{\circ}$ and 60$^{\circ}$) and competitive results in wider FOV (90$^{\circ}$), indicating robust generalization.
In comparison, configurations such as $t=30$, $t=40$, and $t=45$ display more variability and certain limitations in different settings.
Furthermore, the PCA visualization in~\cref{fig:pca} reveals that $t=35$ achieves a favorable balance between noise suppression and spatial information retention, offering latent representations that effectively preserve spatial coherence.
Overall, $t=35$ emerges as a reasonable default choice, offering balanced and reliable performance across a wide range of FOV conditions.

\textbf{Hyperparameter $\lambda$} controls the proportion of the second term in the motion supervision loss.
Specifically, we evaluate $\lambda$ values of 0, 0.05, 0.1, 0.15, and 0.2. 
When $\lambda = 0$, the second term of the motion supervision loss is removed, leading to a considerable decrease in IF, FID and sFID, as shown in~\cref{fig:lam}, which highlights the importance of this loss term in improving the model's performance. 
When $\lambda > 0$, the performance generally improves, and is not sensitive to $\lambda$. 
The performance fluctuations remain small, demonstrating the robustness of our method with respect to $\lambda$. 

By comparing the performance across different FOV settings, we set $\lambda = 0.1$ finally for its best and most stable performance across all conditions. 

\section{Effectiveness of Adaptive Reprojection}
\begin{figure}htb]
    \centering
    \includegraphics[width=\textwidth]{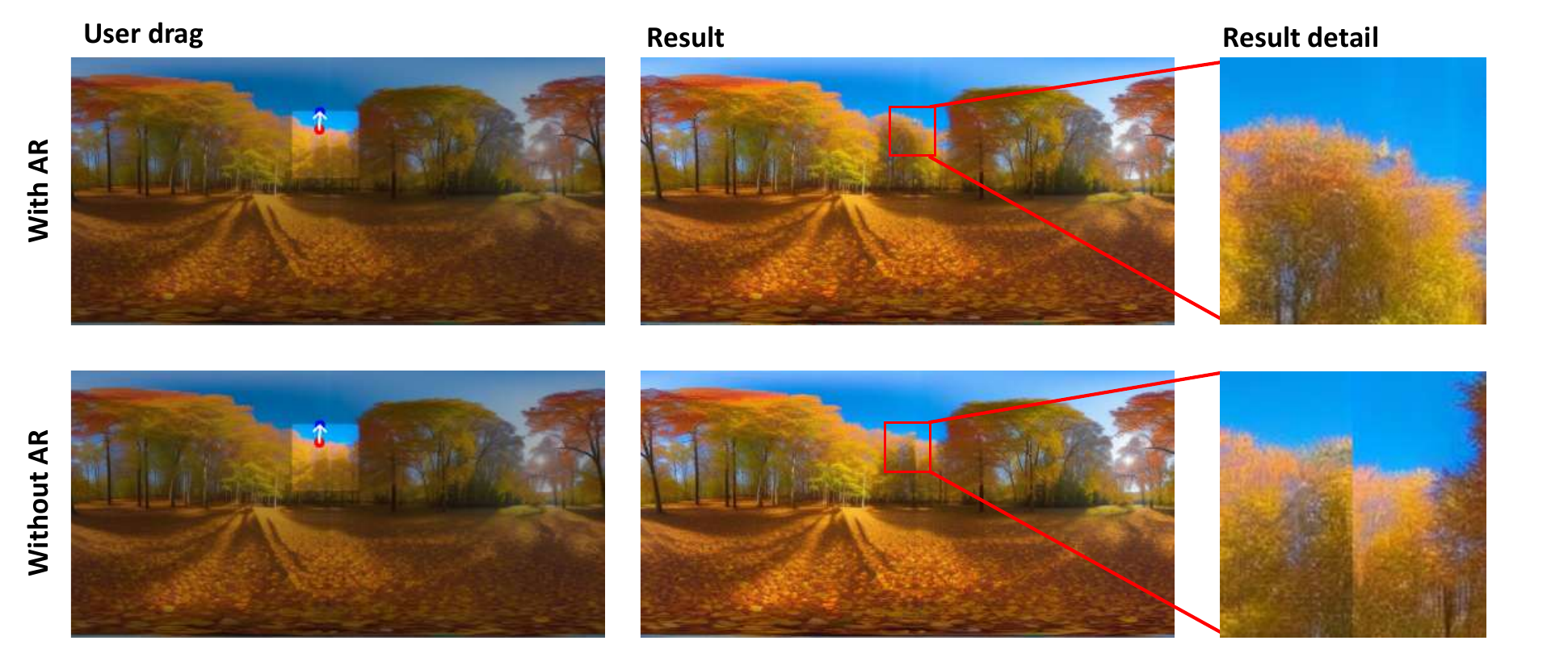}
    \caption{Ablation study on \pointone (\pointones). The top row shows the result with \pointones enabled, while the bottom row shows the result without it. The red boxes highlight the seam regions. With \pointones, the transitions across the seams are smooth and visually coherent. In contrast, without \pointones, noticeable artifacts and discontinuities appear. These results demonstrate the effectiveness of \pointones in enhancing geometric alignment and visual consistency in panoramic image editing.}
    \label{fig:ablation}
\end{figure}

\newpage 

\bibliographystyle{splncs04}
\bibliography{bibli}

\end{document}